\documentclass[review]{elsarticle}
\usepackage{amsmath}  
\usepackage{multirow}
 \usepackage[caption=false]{subfig}
  \usepackage[utf8]{inputenc}
\usepackage{graphicx}
\usepackage{url}
\usepackage{color}
\usepackage{amsmath}
\usepackage{ulem}
\usepackage{multirow}
\usepackage{algorithm}  
\usepackage{algpseudocode}  
\usepackage{amsmath}  

\graphicspath{{MyFigures/}}
\usepackage[caption=false]{subfig}

\usepackage{diagbox} 

\begin{document}

\begin{frontmatter}

\title{RLCFR: Minimize Counterfactual Regret by Deep Reinforcement Learning}





\author[mymainaddress]{Huale Li}
\author[mymainaddress]{Xuan Wang}
\author[mymainaddress]{Fengwei Jia}
\author[mymainaddress]{Yifan Li}
\author[mymainaddress]{Yulin Wu}
\author[mymainaddress]{Jiajia Zhang}
\author[mymainaddress]{Shuhan Qi \corref{mycorrespondingauthor}}
\cortext[mycorrespondingauthor]{Corresponding author}
\ead{shuhanqi@cs.hitsz.edu.cn}

\address[mymainaddress]{Computer Application Research Center, Harbin Institute of Technology, ShenZhen, 518055, ShenZhen, China}

\begin{abstract}
Counterfactual regret minimization (CFR) is a popular method to deal with decision-making problems of two-player zero-sum games with imperfect information. Unlike existing studies that mostly explore for solving larger scale problems or accelerating solution efficiency, we propose a framework, RLCFR, which aims at improving the generalization ability of the CFR method. In the RLCFR, the game strategy is solved by the CFR in a reinforcement learning framework. And the dynamic procedure of iterative interactive strategy updating is modeled as a Markov decision process (MDP). Our method, RLCFR, then learns a policy to select the appropriate way of regret updating in the process of iteration. In addition, a stepwise reward function is formulated to learn the action policy, which is proportional to how well the iteration strategy is at each step. Extensive experimental results on various games have shown that the generalization ability of our method is significantly improved compared with existing state-of-the-art methods.
\end{abstract}

\begin{keyword}
Counterfactual regret minimization\sep decision-making \sep imperfect information \sep reinforcement learning 
\end{keyword}

\end{frontmatter}

\section{Introduction}\label{sec:introduction}
Machine game is one of the most challenging research directions in the field of artificial intelligence, which mainly studies the decision-making problem of players in the environment. According to whether the game state is completely observable, the machine game can be divided into perfect information game (PIG) and imperfect information game (IIG). The PIG means that the game state is completely observable by the game players. On the contrary, the IIG refers to the game in which the players contain private information to other players, in other words, the game state may not be completely observable to the players. For example, in the game of poker, each player's private hand cards are unobservable to other players. Because of such hidden information, in order to solve the IIG, many techniques have been developed to infer or evaluate the hidden information \cite{osborne1994course,Cesa2006Prediction}.

Counterfactual regret minimization (CFR) is one of the most classical method to solve the Nash equilibrium strategy in two-player zero-sum games \cite{zinkevich2008regret,nash1951non,foster1999regret}. There have been many improvements based on the vanilla CFR over the years. For example, Monte carlo CFR (MCCFR) combines the sampling technique monte carlo with the vanilla CFR, which greatly expands scalability of solving problems. CFR+ is a new variant based on the vanilla CFR, which uses the regret matching+ as its core algorithm to speed up the strategy solving \cite{tammelin2015solving,bowling2015heads}. Discount CFR (DCFR) is the latest variant of CFR based method, which obtains the best performance compared with other CFR based methods \cite{brown2019solving}. Deep CFR combines the deep neural network with the LCFR, which approximates the regret value through the neural network, and such combination further expands the scale of solving problems \cite{brown2019deep}. Besides, the CFR based method has also achieved great success in recent years, especially in poker games \cite{johanson2012efficient,burch2013cfr,lisy2015online,brown2015regret}. For example, DeepStack \cite{moravvcik2017deepstack} and Libratus \cite{Brown2017Superhuman} both use the CFR as the core algorithm, which have successfully defeated the top human players in two-player no-limit Texas Hold’em poker. Further, the success of Pluribus that defeats the human professional players in the six-player no-limit Texas Hold’em poker is considered a major breakthrough in the field of artificial intelligence \cite{brown2019superhuman}.

CFR has achieved great success in the IIG, and has many improvement methods over the years. \cite{jackson2016compact,jackson2017targeted,li2018double,steinberger2019single,zhou2018lazy,schmid2019variance}. However, there is still a problem needs to be solved: how to improve the generalization of the CFR based methods. In other words, one method only shows the best performance in one or several games, and no one method can show excellent performance in all games. What is more, for a certain game, a method may only show excellent performance at a certain stage. For example, in the Pluribus \cite{brown2019superhuman}, two different CFR-based methods are set to solve the strategy at different iteration times.
\begin{figure}[!t] 
	\centering  
	\includegraphics[width=\textwidth]{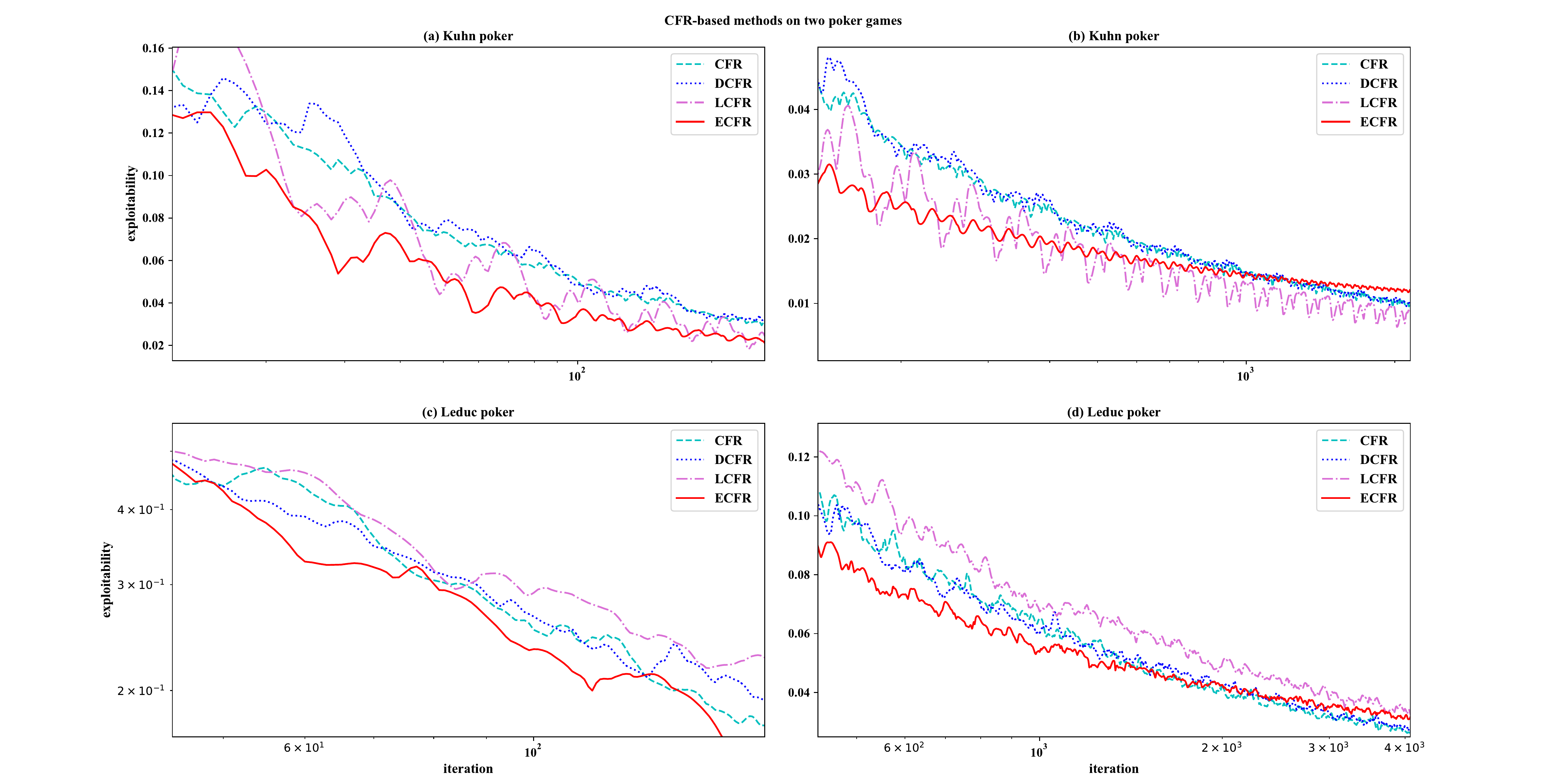}
	\caption{Evaluation of four CFR based methods on Kuhn poker and Leduc poker. (The X-axis represents the number of the iteration, the Y-axis represents the exploitability, which is a classic evaluation metric in the poker games. Here are four curves with different colors, representing four different methods.)
	}
	\label{motivation}  
\end{figure} 

To be specific, as shown in Fig.~\ref{motivation}, we test four CFR based methods in two kinds of poker games. The detail of these two poker games and the Y-axis exploitability will be presented in the section \ref{sec:results} (The exploitability of a strategy in a two-player zero-sum game is how much worse it dose versus a best response compared to a Nash equilibrium strategy. The lower the exploitability, the better the strategy). Here, we just need to know that the smaller of the ordinate is, the better the performance of the method can reach. We can find that in the Kuhn poker, as shown in the Fig.~\ref{motivation}(a) and Fig.~\ref{motivation}(b), the ECFR is better than other three mehtods in the first 50th iterations. The ECFR and the LCFR \cite{brown2019deep} have significant advantages between 50th iterations and 1000th iterations (for example, in the iteration of 108, 190, the LCFR performs better than the ECFR. In the iteration of 60-70, 100, the ECFR performs better). After the 1000th iterations, the LCFR performs the best. In the Leduc poker, as shown in the Fig.~\ref{motivation}(c) and Fig.~\ref{motivation}(d), the ECFR performs better than other three methods in the early 1500 iterations. And the DCFR \cite{brown2019solving} and CFR perform better after 1800th iterations. Thus, it's clear that no one of the CFR variant has obtained completely dominate strategy in the two poker games, and such results further suggest that the existing CFR method has the problem of poor generalization limitation.

To tackle the above problem, different from the traditional CFR method, which fixs the way of regret updating, we adopt a dynamic adaptive method to select different regret updating methods. An agent is designed to dynamically select the most appropriate regret updating method for different stages of strategy iteration. In this way, the selection of regret updating method is always optimal in the whole process of iteratively solving strategy, which improves the generalization ability. Specifically, in the proposed RLCFR, the reinforcement learning method (RL) is adopted to combined with the CFR. In addition, the dynamic process of iterative interactive strategy updating can be modeled as a Markov decision process (MDP), and several regret updating methods can be regared as the actions of the RLCFR. And, at each step, the RLCFR needs to decide the action, which refers to the most appropriate of regret updating method. The above process will be repeated until the maximum number of interactions is reached.

Experimental results indicate that the proposed RLCFR is robust to three different poker games. Given the various poker games, our proposed method shows better generalization ability compared with the state-of-the-art methods. We summarize our contributions as follows:
\begin{itemize}
	\item We put forward a framework, RLCFR, which combines the RL and the CFR. In the framework, the dynamic process of iterative interactive strategy updating is modeled as a MDP. Unlike most of the existing CFR based methods that are proposed for solving larger scale problems or accelerating solution efficiency, we study on how to improve the generalization ability of CFR based methods.
	
	\item We present a new attempt to deal with solving game strategy based on the CFR in a reinforcement learning framework. Based on our method, RLCFR, a policy is learned by the agent to select the appropriate regret updating method, which progressively improves the iteration strategy. Moreover, in order to achieve better performance, a stepwise reward function is formulated to learn the action policy, which is proportional to how well the iteration strategy is at each step.
	
	\item Extensive experimental results show that the generalization ability of our method RLCFR is significantly improved on various poker games, compared with state-of-the-art methods.
\end{itemize}

The paper is organized as follows: the overall framework RLCFR and its details are discussed in Section~\ref{sec:method}. Then, Section~\ref{sec:results} shows experimental results including performance comparsion with the state-of-the-art methods and ablation studies. Finally, the conclusion of this paper is given in Section~\ref{sec:conclusion}.


\section{Methodology}
\label{sec:method}
In this section, the dynamic process of iterative interactive strategy updating is formulated as a MDP firstly. Then, the overall framework of the proposed RLCFR is illuminated. In addition, the proof of convergence for RLCFR is given. Finally, the network and training will be presented in details. 

\subsection{Problem Definition and Overview of RLCFR}\label{sec:_framework}
\subsubsection{Problem Definition}
Given an imperfect information game with two-player, the goal of this paper is to solve a robust strategy by CFR based methods. Meanwhile, we hope that this method has a good generalization ability. In addition, for CFR, the core of it is to solve the problem iteratively through the regret matching algorithm, which means that the strategy of the next iteration is updated by the regret value of the current strategy. The iterative updating process of the game $G$ can be formulated as:
\begin{equation}
\sigma_{t} = U(G,\sigma_1);  \ \ \  U_t = U_1 \circ U_2 \circ \cdots \circ U_t
\end{equation}
where $G$ is the game, $\sigma_t$ is the strategy for the iteration of $t$, $t$ is the number of the iteration. $\circ$ represents the function composition and $U_1, U_2, \dots, U_t$ stands for the specific type of the strategy updating respectively. The ultimate goal is to get the optimal strategy $\sigma_t$ under the same conditions, so that the exploitability of the strategy $\sigma_t$ is minimized.

The procedure of solving the game strategy is a iterative process with two steps. The first step is to obtain the regret of the current strategy, which is calculated by conducting the current strategy. The second step is to obtain the next iteration strategy by the regret matching. This process is repeated until the termination condition is reached, such as the number of iteration is reaching the setting value. In the existing methods, the strategy updating method is fixed in the whole iterative process. From the Fig.~\ref{motivation}, it is already shows that the existing CFR methods \cite{brown2015regret,brown2019deep,brown2019solving,zinkevich2008regret} has the problem of poor generalization limitation.

\begin{figure}[!t] 
	\centering  
	\includegraphics[width=\textwidth]{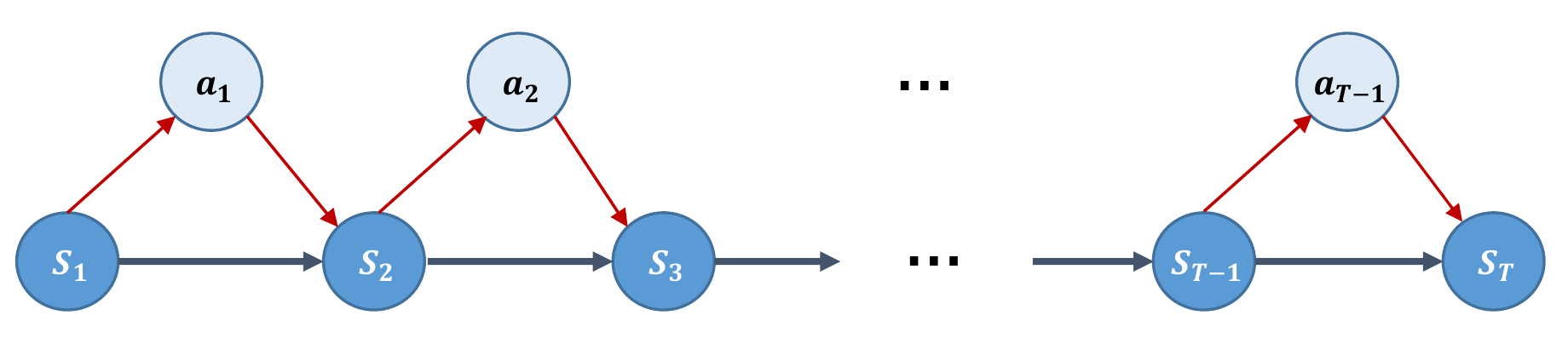}
	\caption{A Markov decision process for the iterative updating process. (In the game $G$, $S_t(\sigma_t, \tilde{R}_t )$, $\sigma_t$ is the strategy and $\tilde{R}_t$ is the regret of the $\sigma_t$. The action $a_t$ is a specific way to update the strategy.)
	}
	\label{mdp}  
\end{figure}

In this case, we assume that the way of strategy updating is critical to the obtained strategy. To address this challenge, we treat the sequential selection of the way of regret calculation and strategy updating problem as a Markov Decision Process (MDP), as shown in the Fig~\ref{mdp}. Meanwhile, we solve it in a deep reinforcement learing manner. The overview of the proposed framework RLCFR (Fig~\ref{framework} ) is provided in the following. 

\subsubsection{Overview of RLCFR} 
\begin{figure}[!t] 
	\centering  
	\includegraphics[width=\textwidth]{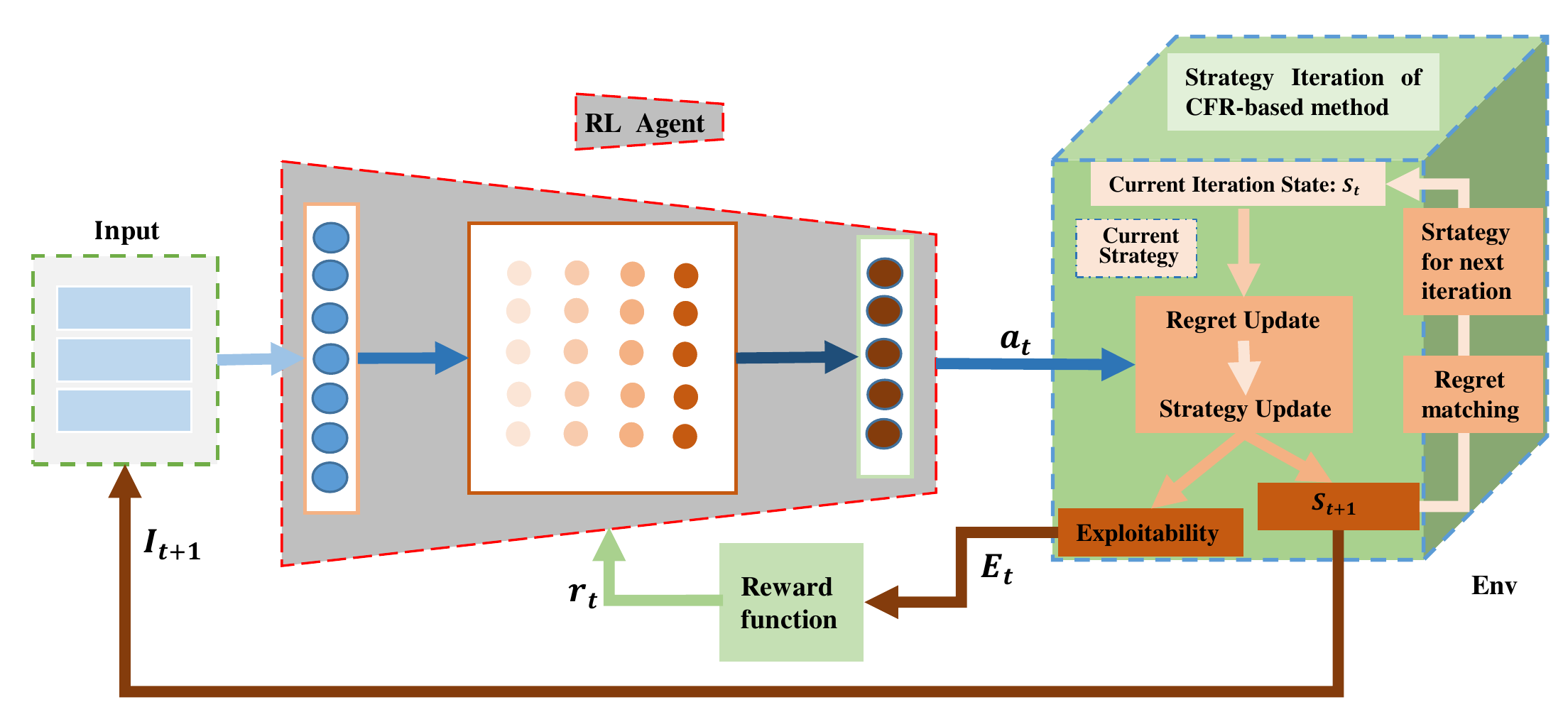}
	\caption{Illustration of our RLCFR framework. (At each step $t$, the RL agent gets the current input $I_t$ from Env, including the regret for each player under the current strategy, which is the output of the agent at each previous step. In order to maximize the reward $r_t$, an action $a_t$ is selected by the agent. After the whole procedure of the regret matching, RLCFR conducts another step until the max step is reached.)
	}
	\label{framework}  
\end{figure}

The proposed framework aims to improve the generalization ability of the CFR by selecting different types of regret calculation and strategy updating. As shown in Fig~\ref{framework}, the proposed RLCFR mainly consists of two componenets, environment and agent: 

(1) Environment (Env). We regard the specific game as the environment, whcih receives the action selected by the agent. At each step, the Env obtains the action provided by the agent and makes the last updating strategy as the current strategy. Then calculates the regret and updates the stratefy of next iteration. Finally, the Env outputs the exploitability of the current strategy, the next iteration strategy and the current strategy's regret. 

(2) Agent. RL agent, which is trained by the reinforcement learning method to select the appropriate types of regret calculation and strategy updating in every step. The agent can be learned from the input and the reward (the input is from the Env and the reward is from the reward function). Then the agent outputs the action to the Env. 

The whole interaction process is that the RL agent get rewards through continuous interaction with the Env, and strategies in the Env are updated iteratively until the termination condition is reached.
\subsection{The component of RLCFR}\label{sec:agent}
The process of RLCFR has been introduced before in the Section~\ref{sec:_framework}. In this section, we will mainly describe the agent, such as action, state and reward, and then go into the details of the agent structure and the strategy iteration procedure.

\textbf{Action.} Action is one of the main components of the agent. Assume that action is represented by $a$, the action space $A$, which is composed of all possible legal actions that the agent can adopt that $A = (a_1, a_2, \dots, a_n)$. When the RL agent performs a step, the agent selects an action $a_t$ and applies it to the Env. At this time, according to the current iteration strategy $\sigma_t$, the regret matching algorithm calculates the regret value and the next iteration strategy $\sigma_{t+1}$ according to the selected action $a_t$. Specifically, different actions can be defined as different ways of regret calculation and strategy updating. In this paper, we use seven types of regret calculation and strategy updating. In other words, the action space consists of seven actions, that is $A = (a_1, a_2, a_3, a_4,a_5, a_6, a_7)$. These actions come from the CFR based methods in recent years. Details are described in Tab.~\ref{action}.
\begin{table*}[!t]
	\caption{The detail of seven actions}
	\centering
	\label{action}       
	\begin{tabular}{c|l}
		\hline
		\textbf{Action}&	\textbf{Strategy updating and regret calculation}	\\
		\hline \hline
		$a_1$ &	$
		\sigma_{i}^{T+1}(I,a)=\frac{R_{i}^{T,+}(I, a)}{\sum_{a \in A(I)} R_{i}^{T,+}\left(I, a \right)}, 
		\ \ \	R_{i}^{T,+}(I, a) = \sum_{a \in A(I)} r(I,a)
		$\\
		$a_2$&	$
		\sigma_{i}^{T+1}(I,a)=\frac{t R_{i}^{T,+}(I, a)}{\sum_{a \in A(I)} t R_{i}^{T,+}\left(I, a \right)} , \ \ \
		R_{i}^{T,+}(I, a) = \sum_{a \in A(I)} t r(I,a)
		$\\
		$a_3$&	$
		\sigma_{i}^{T+1}(I,a)=\frac{\left(\frac{t}{t+1}\right)^{\gamma} R_{i}^{T}(I, a)}{\sum_{a \in A(I)} \left(\frac{t}{t+1}\right)^{\gamma} R_{i}^{T}\left(I, a \right)}, \ 	R_{i}^{T}(I, a) = \sum_{a \in A(I)} \frac{t^{\alpha}}{t^{\alpha}+1} r(I,a)
		$\\
		$a_4$&	$
		\sigma_{i}^{T+1}(I,a)=\frac{e^\alpha R_{i}^{T}(I, a)}{\sum_{a \in A(I)} R_{i}^{T}\left(I, a \right)}, \ \ \
		R_{i}^{T}(I, a) = \sum_{a \in A(I)}e^\alpha r(I,a)
		$\\
		$a_5$&	$
		\sigma_{i}^{T+1}(I,a)=\frac{R_{i}^{T}(I, a)}{\sum_{a \in A(I)} R_{i}^{T}\left(I, a \right)}, \ \ \
		R_{i}^{T}(I, a) = \sum_{a \in A(I)} r(I,a)
		$\\
		$a_6$&	$
		\sigma_{i}^{T+1}(I,a)=\frac{R_{i}^{T}(I, a)}{\sum_{a \in A(I)} R_{i}^{T}\left(I, a \right)}, \ \ \
		R_{i}^{T}(I, a) = \sum_{a \in A(I)} r^{+}(I,a)
		$\\
		$a_7$&	$
		\sigma_{i}^{T+1}(I,a)=\frac{1}{|A|}, \ \ \
		R_{i}^{T}(I, a) = \sum_{a \in A(I)} r(I,a)
		$\\
		\hline
	\end{tabular}
\end{table*}

\textbf{State.} State is the observation information of the agent obtained from environment under the current condition. In the RLCFR, we take the regret value of the current strategy as the output $S_t$, and the input of the agent $I_t$ is the vector of the regret value of the last three iterations, which is a three-dimensional vector that $I_t = (S_{t-2}, S_{t-1}, S_{t})$. At the step 1, $I_1$ can be initialized randomly. 

\textbf{Reward.} Reward is a basic element of reinforcement learning method. Appropriate reward setting can not only accelerate the convergence of the training model, but also make the trained agent more robust. In addition, duo to the exploitability is a very objective evaluation metric of the current iteration strategy, the lower of the exploitability is, the better of the current strategy is. Therefore, we hope that the reward can be associated with at each step. Thus, a stepwise reward is designed as follows:
\begin{equation}
R_2=\left\{\begin{array}{l}
E_{t}^{'} - E_t, \ \ \ \ \ \  if \ \ \ E_{t}^{'} - E_t \geq 0 \\
E_{t}^{'} - E_t, \ \ \ \ \  \ otherwise
\end{array}\right.
\label{R2}
\end{equation}
where $E_{t}^{'}$ is the lowest exploitability of the comparison methods under the same number of iterations.

In general, the accumulated reward of one interactive sequence is as follows:
\begin{equation}
R_i = \Sigma _{t=1}^T \gamma^{t-1} r_i^t
\label{R}
\end{equation}
where $T$ is the number of the total step, $\gamma$ is the discount factor and its value is between 0 and 1. 

\textbf{Env.} In the reinforcement learning, the agent is improveed by constantly interacting with the environment. It is worth noting that in RLCFR, although the Env here is built by ourselves, it is not different from the general environment, and also includes the same elements of the environment. The Env receives the action output by the agent for execution, and generates the next state and corresponding rewards. Specially, the Env receives the action output from the agent and executes the current strategy $\sigma_t$ in the IIG. Then the way of regret updating and regret calculation is selected through the received action for the regret matching algorithm. Finally, the Env outputs the exploitability $E_t$ and the regret $\tilde{R}_t$ of the current strategy, and the updated strategy $\sigma_{t+1}$ for the next iteration.

\subsection{Proof of The Convergence}
In this section we prove a bound for the proposed approach RLCFR when calculating the average strategy. Theorem 1 shows the RLCFR has a convergence bound. 

\textbf{Theorem 1}  Assume that the number of the iteration $T$ of the RLCFR, which is played in a two-player zero-sum game. Then the weighted average strategy profile is a 
$2\frac{\Delta |\mathcal{I}| \sqrt{|A|}} {\sqrt{T}}$-Nash equilibrium.

The proof is provided in the appendix, which combines the the proof for the vanilla CFR \cite{zinkevich2008regret}, CFR+ \cite{tammelin2015solving,bowling2015heads} and the discount CFR \cite{brown2019solving}.

\subsection{Training of RLCFR}\label{sec:_training}
In this section, the agent is trained based on the RL. In RLCFR, the simplified version of the deep Q-larning (DQN) is used to train the agent. Compared with the vanilla DQN \cite{mnih2015human}, there is no complex input similar to the image in RLCFR, thus we remove the convolutional neural network and only use full connection to construct the network of the RLCFR. In addition, like the original DQN, we also use the target network. The loss function in RLCFR is a mean square error (MSE),
\begin{equation}
L = \frac{1}{2}||y-y^{'}||_2^2
\end{equation}

As for the agent, part of the supervision information we used when setting up the reward function to train the agent (the minimum exploitability of each CFR method at the same number of iterations). And in the training procedure of the agent, the goal is to minimize the loss function. In addition, the loss function of the agent is the same as the vanilla DQN. The loss function $L$ can be rewrited with:
\begin{equation}
L = (y_t - Q(s_t,a))^2 
\end{equation}
the $y_t$ is as follows:
\begin{equation}
y_t = \left\{\begin{array}{l}
{r_t + \gamma max_{a^{'}}Q_{t+1,a^{'}} } \ ,\ \ \ \ \  T > t \geq 1 \\
{r_T} \ , \ \ \ \ \ \ \ \ \ \ \ \ \ \ \ \ \ \ \ \ \ \ \ \ \ \ \  t = T
\end{array}\right.
\end{equation} 
where $Q_a$ is the action value of the action $a$, $\gamma$ is a discount factor and the value is set to 0.99.

\section{Experiments}\label{sec:results}

In this section, we conduct experiments to evaluate the performance of our proposed RLCFR in terms of the exploitability, as compared  with the state-of-the-art methods. In addition, we investigate different settings of the proposed RLCFR.

\subsection{Experimental Platform}
We perform experiments on the poker games, which are the most classic and universal testing platform in the field of the IIG in recent years. Since the poker game contains private hand and includes all the elements of the IIG, it can well evaluate the method of the IIG. In addition, in recent years, the successful IIG methods (for example, DeepStack \cite{moravvcik2017deepstack}, Libratus \cite{Brown2017Superhuman}, Pluribus \cite{brown2019superhuman}) adopt the poker game as the test platform for experimental testing. Therefore, in this paper, we also take poker games as our experimental platform. Specifically, three kinds of poker games are used as test platforms in the paper. They are all simple versions of Texas Hold'em poker, which are enough to verify the effectiveness of our method.

The three kinds of poker games are Kuhn poker, Leduc poker, Royal poker respectively. The Kuhn poker has two game players and only one round, preflop. Each game player has one private hand and the game has three cards totally. The Leduc poker has two game players and two rounds, preflop and flop. Each game player has one hand and the game has six cards totally. Each game player is issued a private hand firstly in the round of preflop, and then a public card will be issued in the round of flop. The Royal poker also has two game players and three rounds, preflop, flop, and turn. Each game player has one private hand and the game has eight cards totally. Each game player is issued a private hand firstly in the round of preflop, and then a public card will be issued in the round of flop and turn respectively. Besides, the actions fold, call, and raise are legal actions in the poker games. The detail of these three kinds of poker games are introduced in the Tab.~\ref{poker}.

\begin{table*}[!t]
	\caption{The detail of three kinds of poker}
	\centering
	\label{poker}       
	\begin{tabular}{|c|c|c|c|c|c|}
		\hline
		\textbf{Poker}&	\textbf{Total cards}&	\textbf{Public/private cards}&	\textbf{Round}&	\textbf{Ante}&	\textbf{Betsize} \\
		\hline \hline
		Kuhn&	3&	0/1&	1	&1&	1\\ \hline
		Leduc&	6&	1/1&	2&	1&	2; 4\\ \hline
		Royal&	8&	2/1&	3&	1&	2; 4; 4\\ 
		\hline
	\end{tabular}
\end{table*}

\subsection{Experimental Settings}
\textbf{Evaluation metrics}. Exploitability is a natural and standard evaluation metric in the field of the IIG. In this paper, the exploitability is used to evaluate the strategy solved by the method. The exploitability of a strategy in a two-player zero-sum game is how much worse it dose versus a best response compared to a Nash equilibrium strategy. The lower the exploitability, the better the strategy.

Note that the best response to the strategy $\sigma_p$ is a strategy $Br(\sigma_p)$, in which $u_p(\sigma_p,Br(\sigma_p)) = max_{\sigma^{'}_{-p}} u_i(\sigma_p,\sigma^{'}_{-p})$. A Nash equilibrium strategy $\sigma^*$ is a strategy profile, in which every player plays a best response: $\forall p, u_{p}\left(\sigma_{p}^{*}, \sigma_{-p}^{*}\right)=\max _{\sigma_{p}^{\prime}} u_{p}\left(\sigma_{p}^{\prime}, \sigma_{-p}^{*}\right)$. Thus, the exploitability of a strategy can be descripted as:
\begin{equation}
e\left(\sigma_{p}\right)=u_{p}\left(\sigma_{p}^{*}, B r\left(\sigma_{p}^{*}\right)\right)-u_{p}\left(\sigma_{p}, Br\left(\sigma_{p}\right)\right)
\end{equation}

\textbf{Implementation details.} Our method RLCFR is combining the DRL and CFR based methods to improve generalization ability. Specifically, we use DQN for action selection, which we call it as dqn1 \cite{mnih2015human}. Among them, the dqn1 is slightly different from the vanilla DQN presented by DeepMind. The biggest difference is that the dqn1 only use a four-layer fully connected network to achieve our goal, which does not apply the convolutional neural network. Due to a fully connected network is enough to encode the state in the RLCFR. Besides, other parameters in the dqn1 are: batch size is 32, discount factor is 0.99, replay memory size is 2000, exploration is 0.1. In addition, the target network updates for every 200 steps. The model training and testing on a PC with i7-7700 CPU, 16G RAM. The program language used is python 3.6 and pytorch.
\subsection{Experimental Results}
In order to verify the effectiveness of our method RLCFR, we carried out two groups of experiments totally. Firstly, we compared with the state-of-the-art methods in the field of the IIG, which is in the first group of experiment. Secondly, we conducted ablation studies, including the reward function setting and the action setting. The rest experiments are all about ablation experiments except the first group experiment. In addition, in order to ensure the fairness of the experimental results and reduce the errors caused by the randomness. Thus we have carried out each group of experiments three times respectively and the final experimental results take the mean value.

\subsubsection{Comparisons with state-of-the-art methods} 
We compare RLCFR with four state-of-the-art methods: CFR \cite{zinkevich2008regret}, LCFR \cite{brown2019deep}, DCFR \cite{brown2019solving}, ECFR. Among them, the DCFR is the CFR based method with the best performance  that has been published. In addition, the performance of the ECFR is also very excellent. However, this is our recent work and it has not yet officially published. Thus here we still describe the DCFR as the method with best performance at present. The brief description of the these four comparison methods are in Tab.~\ref{tab:2}.

\begin{table*}[!t]
	\caption{The brief description of the four comparison methods}
	
	\label{tab:2}       
	\begin{tabular}{cl}
		\hline
		\textbf{Method}&	\textbf{Brief description}\\
		\hline \hline
		CFR&	It is proposed by Zinkevich et al. in 2007 \cite{zinkevich2008regret}, which is the first \\& method to use regret matching to solve the poker games.\\ \hline
		LCFR&	It is proposed by Noam Brown et al. in 2019 \cite{brown2019deep}, which is identical \\&to the CFR, except the regret is weighed by the iteration.\\ \hline
		DCFR&	It is proposed by Noam Brown et al. in 2019 \cite{brown2019solving}, which is a variant\\&of the CFR that discount prior iterations, leading to stronger\\& performance than the prior state-of-the-art methods.\\ \hline
		ECFR&	It is also based on the vanilla CFR method and is proposed by us. \\&The ECFR also shows excellent performance through exponential \\&decay of the CFR.\\ 
		\hline
	\end{tabular}
\end{table*}

\begin{figure}[!t] 
	\centering  
	\includegraphics[width=\textwidth]{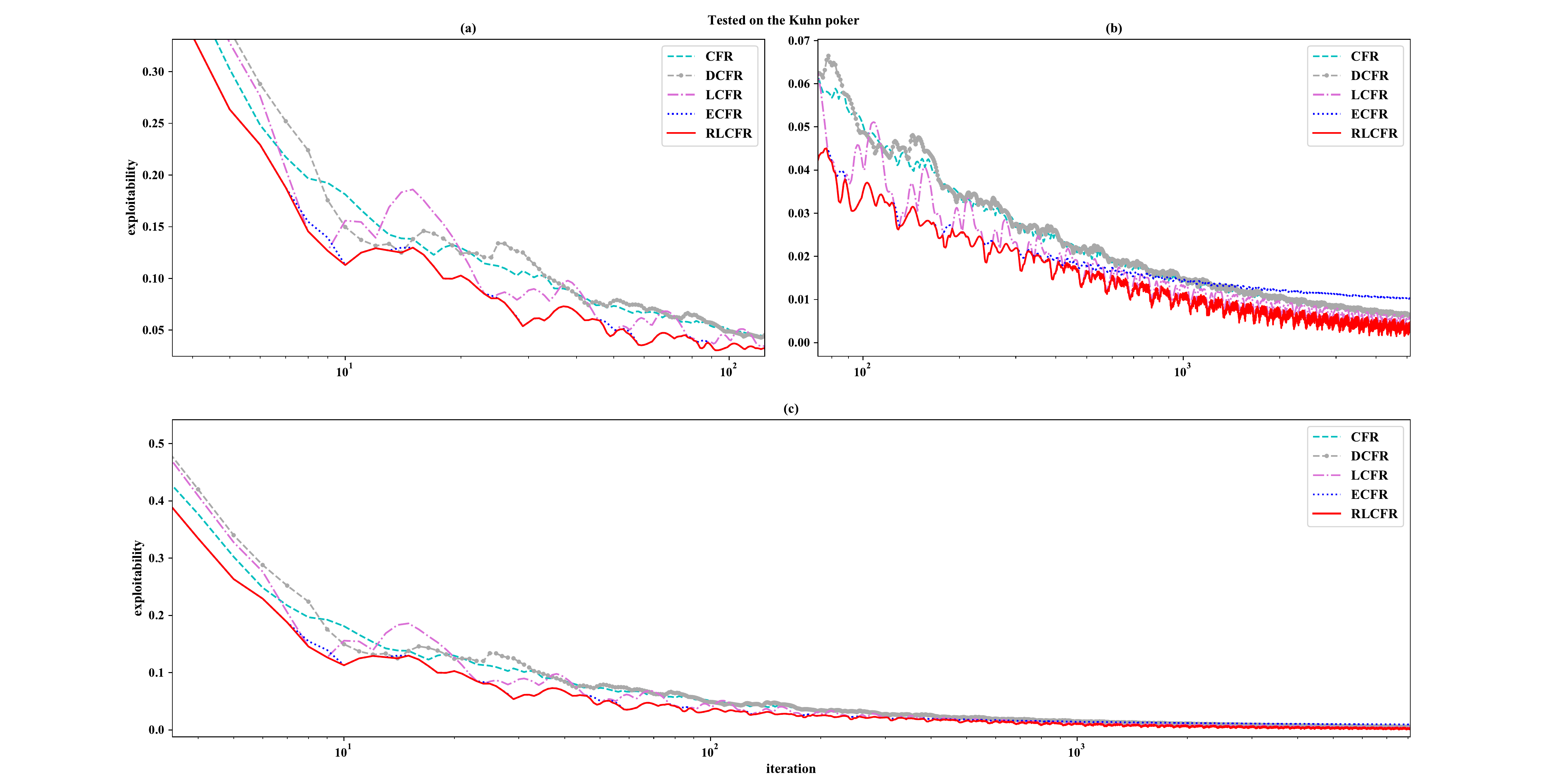}
	\caption{Comparisons with state-of-the-art methods on the Kuhn. (The X-axis represents the number of the iteration, the Y-axis represents the exploitability that is a classic evaluation metric in the poker games. The smaller of the exploitability, the better. Here are five curves of different colors, representing five different methods. Our mehtod RL-CFR is represented with a solid red line. The other four comparison mehtods are represented with dashed lines with different colors.)
	}
	\label{kuhn}  
\end{figure}

\begin{figure}[!t] 
	\centering  
	\includegraphics[width=\textwidth]{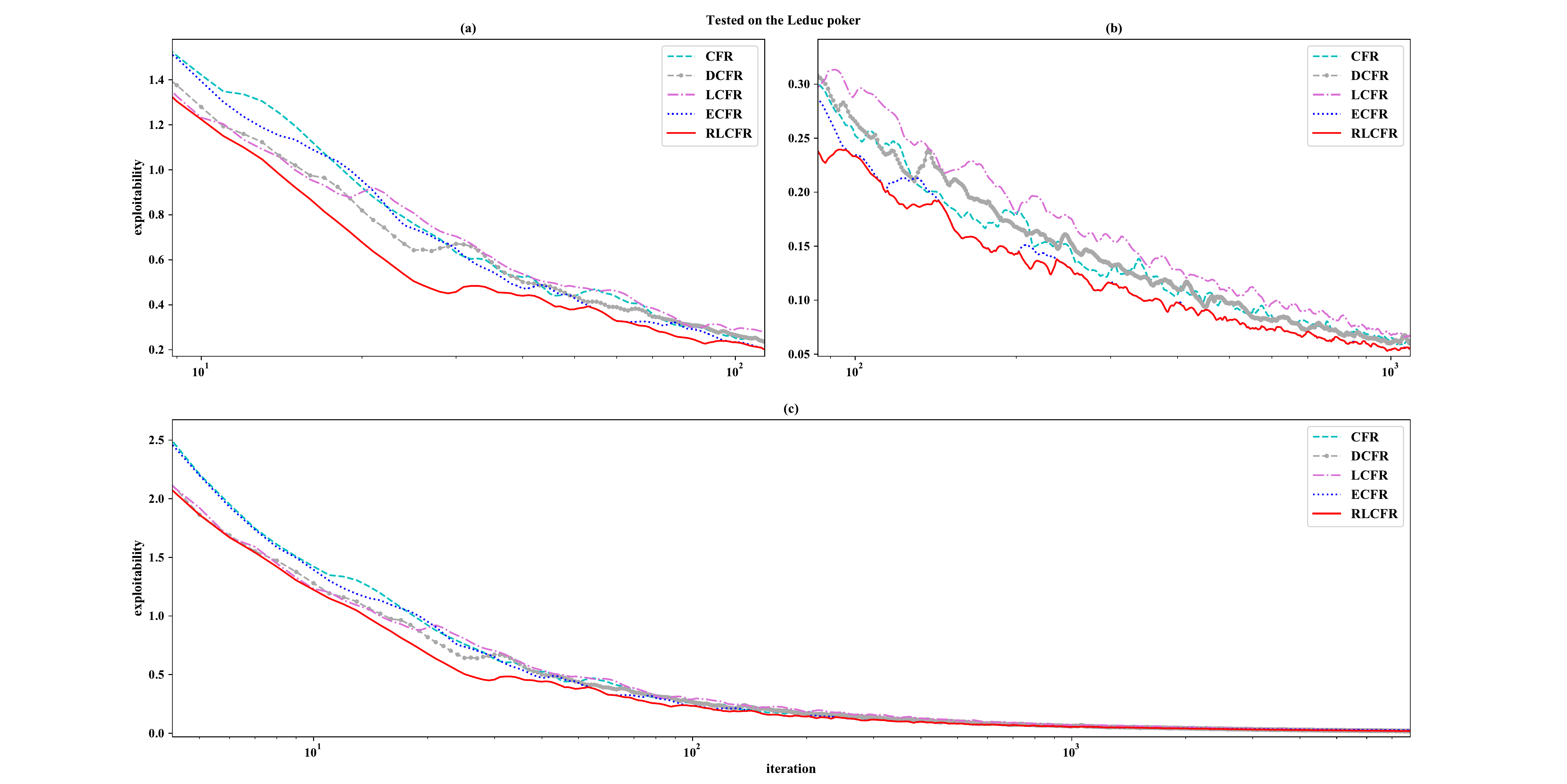}
	\caption{Comparisons with state-of-the-art methods on the Leduc. (The X-axis represents the number of the iteration, the Y-axis represents the exploitability. The smaller of the exploitability, the better.)}
	\label{leduc}  
\end{figure}

\begin{figure}[!t] 
	\centering  
	\includegraphics[width=\textwidth]{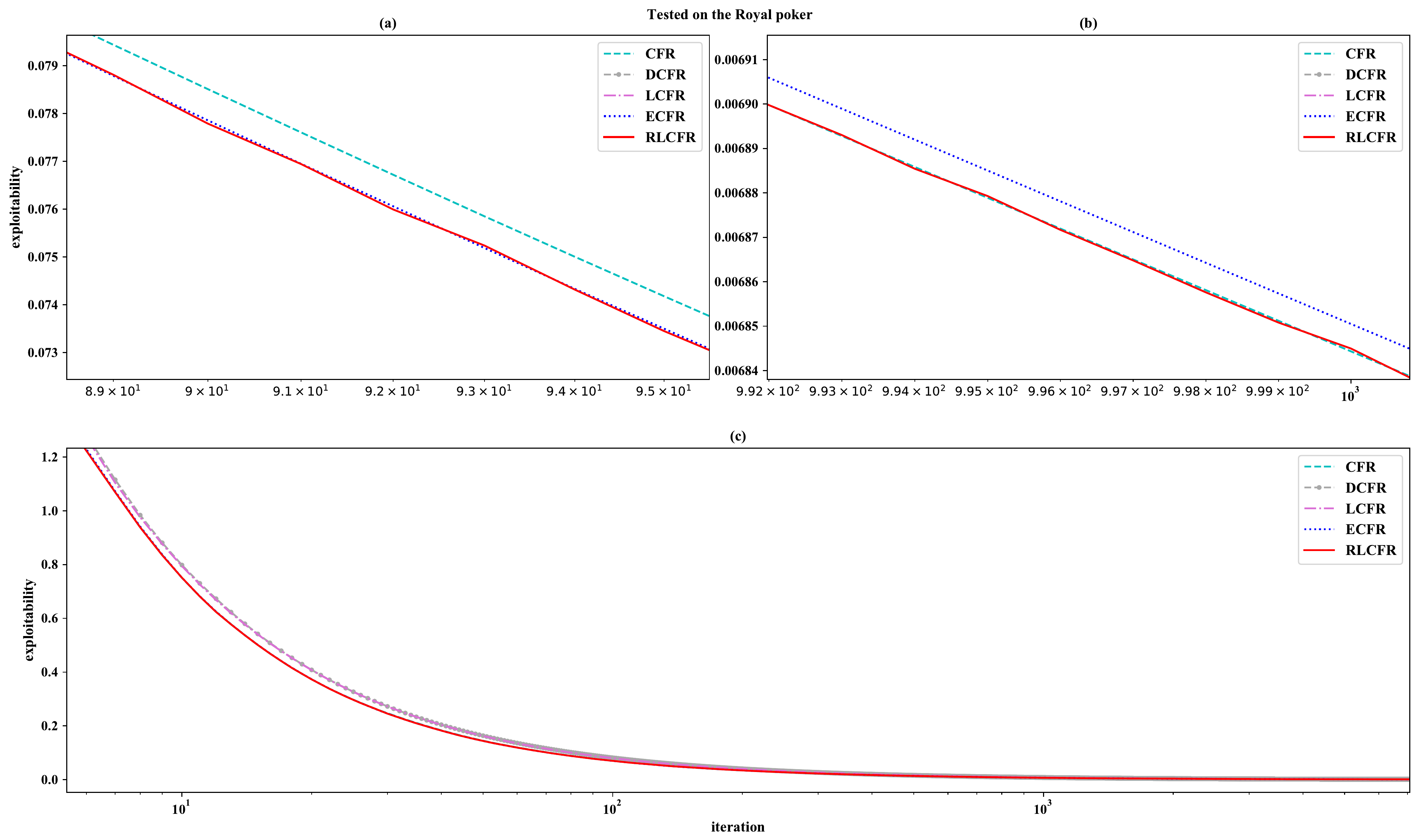}
	\caption{Comparisons with state-of-the-art methods on the Royal. (The X-axis represents the number of the iteration, the Y-axis represents the exploitability. The smaller of the exploitability, the better.)	}
	\label{royal}  
\end{figure}

In order to make the experimental results more fair, we carried out three verification experiments on three kinds of poker games. The final results come from the mean of three validation experiments. In addition, the number of iterations of the strategy is 10000 in the validation experiment. The model selected in the validation experiment is an optimal model after the ablation study, which will be described later. The results are shown in Fig.~\ref{kuhn}, Fig.~\ref{leduc} and Fig.~\ref{royal}, which are tested on the Kuhn poker, Leduc poker and Royal poker respectively. The lower the exploitability, the better the strategy.

In the Fig.~\ref{kuhn}(c), we can find that the RLCFR converges with the iteration increases, which shows that our method has good convergence on the Kuhn poker. Fig.~\ref{kuhn}(a) and Fig.~\ref{kuhn}(b) shows the detail of the experimental results. In the first 100th iterations, the performance of the RLCFR is very close to that of the ECFR, except for a limited number of iterations (at $n$ = 8, 9, 50, 85). After 100th iterations, the performance of the RLCFR is gradually better than that of the ECFR, and it is obviously better than other methods.

In the Fig.~\ref{leduc}(c), we can also find that the RLCFR converges gradually with the increase of iterations. Besides, in addition to a limited number of iterations, the performance is equal to that of ECFR, and we can clearly see that the performance of the RLCFR is ahead of other methods in the iteration process. The same convergence can still be obtained from the Fig.~\ref{royal}(c). We find that in the first 200th iterations, as shown in Fig.~\ref{royal}(a),  the performance of the RLCFR is close to that of the ECFR and there is no dominant trend. However, in the later iterations, especially after 900th iterations, we can obviously find that our method RLCFR is better than the ECFR, which is shown in Fig.~\ref{royal}(b).

To sum up, firstly, our method RLCFR has a good convergence with the increase of the iterations, which has been verified in the three experiments on three different poker games respectively. Secondly, the generalization of our method has been well verified, and our method has shown nearly leading performance in these three different poker games. In addition, our model is trained on Kuhn poker and Leduc poker, while the test is conducted on three kinds of poker games and the additional Royal poker has the largest scale among the three kinds of Poker games. However, our method still shows excellent performance in the Royal poker, which further proves that our method has good generalization ability. Finally, from the Fig.~\ref{kuhn}, Fig.~\ref{leduc} and Fig.~\ref{royal}, we can find that when the performance of several methods is alternately dominant, our method does not oscillate like a single method. On the contrary, our method can keep the optimal or nearly optimal in this kind of situation. This shows that our method is effective in the selection of strategy updating in the iterative process.

\subsubsection{Ablation studies} 
In this section, we investigate different settings of the proposed RLCFR, and give some insights on the choice of each factor. There are two groups experiments to analyze. Firstly, we analyze the effect of different reward settings to the algorithm performance. Secondly, we analyze the effect of different action settings on the performance of the algorithm. It should be noted that the test model we used in the first group of experiments (comparisons with state-of-the-art methods) came from the best combination selected here. In each of the four groups of experiments, we also conducted five repeated experiments to ensure the effectiveness of the experimental results. The final experimental results come from the mean of five repeated experiments.

\textbf{Reward setting}. The reward function is a very important part of the reinforcement learning. An appropriate reward function can accelerate the convergence of the reinforcement learning. Since we also use the deep reinforcement learning method in this paper, it is necessary to set the reward reasonably. In addition, the purpose of using reinforcement learning method is to select an appropriate action through reinforcement learning method, and to select the strategy updating method in CFR based methods. The evaluation standard of the CFR methods is exploitability. Thus, we can find the reward function is very related to the exploitability.  

Due to the CFR is an iterative algorithm, with the increase of iterations, the solution strategy should be stronger and stronger. In terms of the exploitability, with the increase of iterations, the exploitability should be lower and lower. Based on the above analysis, we set up three different rewards, which are called $R_1, R_2, R_3$. 

\textbf{$R_1$:} we compare the exploitability of the two steps before and after, since the lower the exploitability is, the better. Thus when the exploitability develops according to this trend, we will give a positive reward, otherwise it will be a negative reward. As showed in the Eq.~\ref{R1}.
\begin{equation}
R_1=\left\{\begin{array}{l}
1, \ \ \ \ \ \ \ \ \ if \ \ \ E_{t-1} - E_t >0 \\
-1, \ \ \ \ \  \ otherwise
\end{array}\right.
\label{R1}
\end{equation}
where $t$ is the number of the iteration, $E_{t-1}$ and $E_{t}$ is the exploitability of the RLCFR under the previous iteration and the current iteration.

\textbf{$R_2$:} In the previous completed experiments, we found that in most cases, the exploitability decreased gradually. That is to say, according to the $R_1$ setting, the reward is positive in most cases. Due to the punishment is too little, it has influence on the training of the RL model. Moreover, it will cause the model to actually choose risky actions when facing some states, but the model is not considered as risky actions. Therefore, we have improved $R_1$ and added a benchmark to $R_1$, through the exploitability we have obtained from various comparison methods in each iteration. As depicted in the Eq.~\ref{R2}.
\begin{equation}
R_2=\left\{\begin{array}{l}
E_{t}^{'} - E_t, \ \ \ \ \ \  if \ \ \ E_{t}^{'} - E_t \geq 0 \\
E_{t}^{'} - E_t, \ \ \ \ \  \ otherwise
\end{array}\right.
\label{R2}
\end{equation}
where $E_{t}^{'}$ is the lowest exploitability of the comparison methods under the same number of iterations.

\textbf{$R_3$:} At the same time, the attenuation rate of the exploitability is also worth considering. For the iterative method, the number of iterations setting in practical application is limited. The fewer iteration is used to achieve the same exploitability, the more beneficial to the practical application. Therefore, we add an additional bonus item in $R_3$ to encourage corresponding actions that lead to faster attenuation speed. As showed in the Eq.~\ref{R3}.
\begin{equation}
R_3=\left\{\begin{array}{l}
\frac{E_{t}^{'} - E_t}{E_{t}^{'}}, \ \ \ \ \ \ \ \ \ if \ \ \ E_{t}^{'} - E_t \geq 0 \\
\frac{1}{k_{t} - k_{t-1}}, \ \ \ \ \  \ \ \   if \ \ \ k_{t} - k_{t-1} > 0 \\
\frac{E_{t}^{'} - E_t}{E_{t}^{'}}, \ \ \ \ \ \ \ \ \ otherwise
\end{array}\right.
\label{R3}
\end{equation}
where $k_t$ is the slope, which is described as $\frac{E_{t^{'}}-E_{t}}{t-t^{'}}$, here $t^{'} < t$. $E_{t}^{'}$ is the lowest exploitability of the comparison methods under the same number of iterations. $t$ is the number of the iteration, $E_{t-1}$ and $E_{t}$ is the exploitability of the RLCFR under the previous iteration and the current iteration. 

\begin{figure}[!t] 
	\centering  
	\includegraphics[width=\textwidth]{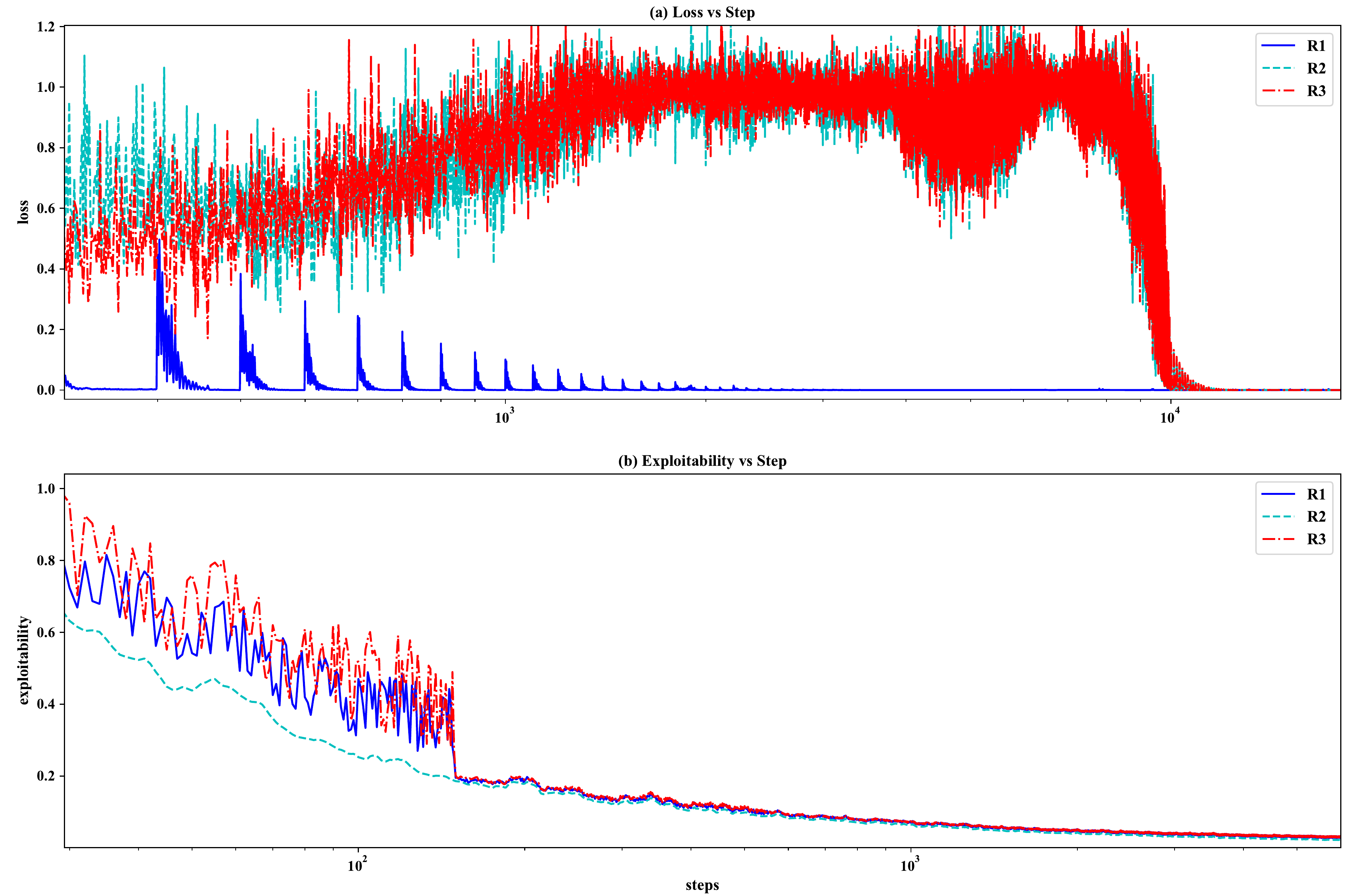}
	\caption{Ablation study on reward functions. (The X-axis represents the number of the step, the Y-axis represents the exploitability. The smaller of the exploitability, the better.)
	}
	\label{fig:reward}  
\end{figure}

In this study, we tested the three kinds of the reward to find an appropriate reward in the RLCFR. The training steps here is 20000. The related experiental results are showed in Fig.~\ref{fig:reward}. The model loss and the exploitability are used to evaluate our results respectively. As can be seen in Fig.~\ref{fig:reward}(a), the model used $R_1$ converges fast compared with the other two. Although the models used $R_2$ and $R_3$ converge finally after the 10000 steps. From the Fig.~\ref{fig:reward}(b), we find that the model used $R_2$ performs the best. In addition, the model of $R_3$ does not show the expected effect. This is mainly due to the rapid decline in exploitability in the early stage, and the use of slope can be beneficial to provide more incentive information. However, in the later iteration, the exploitability decreases slowly, and the attenuation slope decreases all the time. This is equivalent to punishing the agent without rewards, which makes the model training of the agent more difficult. Finally, since our ultimate goal is to reduce the exploitability of the method, we choose the $R_2$ as the reward function. 

\textbf{ Actions.} The action is one of the basic concepts of the RL. Reasonable action settings have a great influence on the RL methods. In our proposed method RLCFR, based on the analysis of CFR methods, we set up four necessary actions. In addition, we added three actions to explore their impact on the RL. The first action is based on the idea of vanilla regret matching algorithm, which is to follow the process of updating regret completely without additional preprocessing. The second action is also from a very direct motivation, which is just to accumulate the positive regret and use the positive regret to update the strategy. It should be noted that this action is different from the action update of the CFR. The CFR only preprocesses the regret when the strategy is updating. The third action is random strategy updating. The consideration of adding this action is to increase the randomness of the action and ultimately increase the flexibility of the training model.

\begin{figure}[!t] 
	\centering  
	\includegraphics[width=1\textwidth]{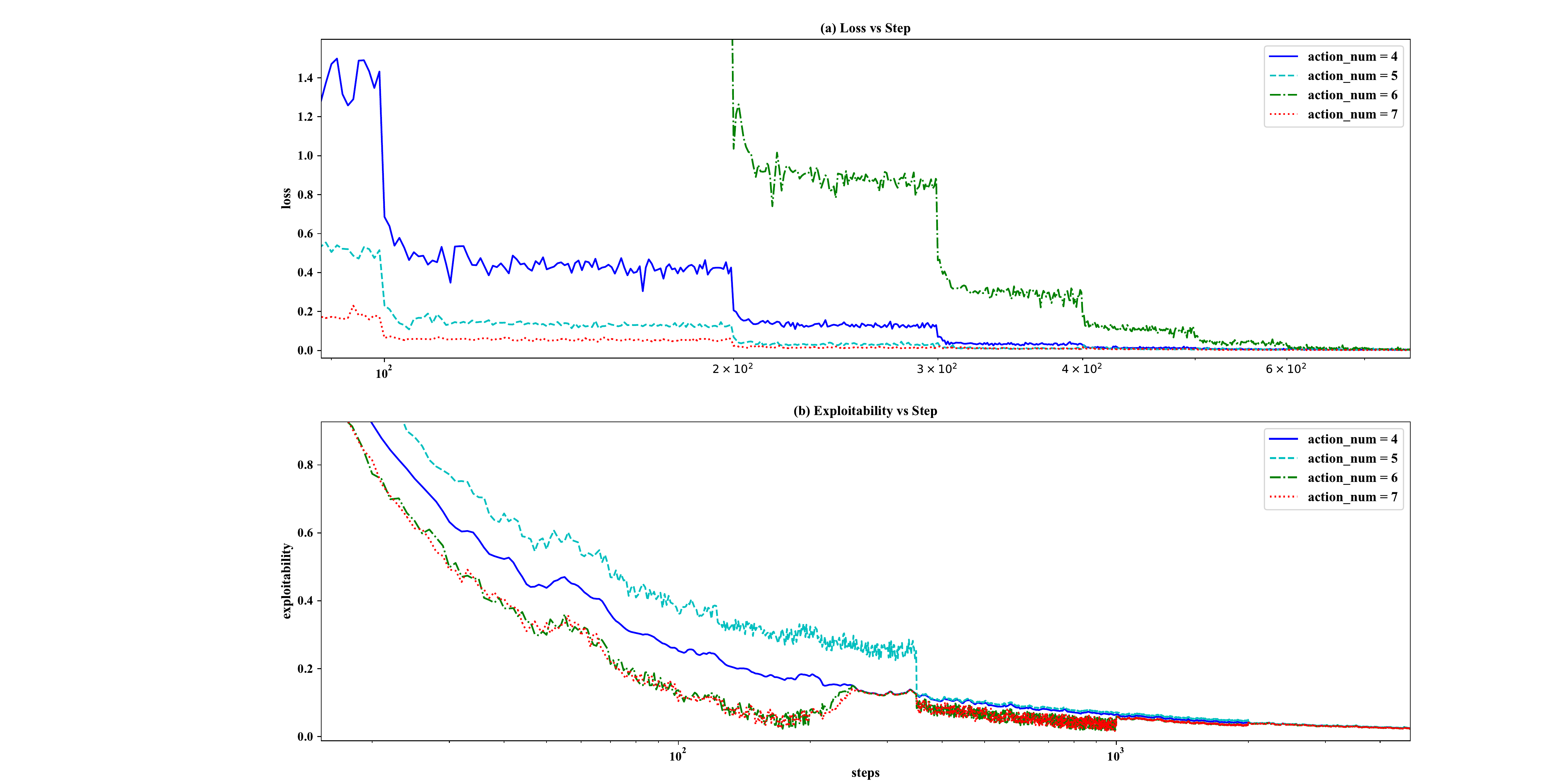}
	\caption{Ablation study on actions. (The X-axis represents the number of the step, the Y-axis represents the exploitability. The smaller of the exploitability, the better.)
	}
	\label{fig:action}  
\end{figure}

We call the three additional actions as $a_5$, $a_6$, $a_7$ respectively. In addition, in order to distinguish the four basic actions, we express the four basic actions as $a_1, a_2, a_3, a_4$. To demonstrate the effectiveness of the actions $a_5$, $a_6$, $a_7$, we compare the results with/without the actions $a_5$, $a_6$, $a_7$. As can be seen in Fig.~\ref{fig:action}, we can find that no matter from the perspective of loss or exploitability, the added actions are helpful to the training model. Although six actions and seven actions are similar in exploitability, the loss of the model using seven actions decreases faster. In addition, seven actions can increase the agent's exploratory ability by randomly adopting updating strategies. Therefore, we finally chose seven actions in the test model.

\section{Conclusions}\label{sec:conclusion}

In this paper, in order to improve the generalization ability of CFR based methods, we put forward a framework, RLCFR, which combines the RL and the CFR. In the framework, the dynamic process of iterative interactive strategy updating is modeled as a Markov decision process. We present a new attempt to deal with solving game strategy based on the CFR in a reinforcement learning framework. Based on our method, RLCFR, a policy is learned by the agent to select the appropriate regret updating method, which progressively improves the iteration strategy. Moreover, in order to achieve better performance, a stepwise reward function is formulated to learn the action policy, which is proportional to how well the iteration strategy at each step. Extensive experimental results on three kinds of games verified the effectiveness of the proposed approach.

\section*{Acknowledgment}
This research is supported by  Key Technology Program of Shenzhen, China, (No.JSGG20170823152809704), Key Technology Program of Shenzhen, China, (No.JSGG20170824163239586), and Basic Research Project of Shenzhen, China, (No.JCYJ20180507183624136).\textsc{}
\section*{References}

{\footnotesize
	\bibliography{biblio}}

\begin{thebibliography}{10}
\expandafter\ifx\csname url\endcsname\relax
  \def\url#1{\texttt{#1}}\fi
\expandafter\ifx\csname urlprefix\endcsname\relax\def\urlprefix{URL }\fi
\expandafter\ifx\csname href\endcsname\relax
  \def\href#1#2{#2} \def\path#1{#1}\fi

\bibitem{osborne1994course}
M.~J. Osborne, A.~Rubinstein, A course in game theory, MIT press, 1994.

\bibitem{Cesa2006Prediction}
N.~Cesa-Bianchi, G.~Lugosi, Prediction, Learning, and Games, 2006.

\bibitem{zinkevich2008regret}
M.~Zinkevich, M.~Johanson, M.~Bowling, C.~Piccione, Regret minimization in
  games with incomplete information, in: Advances in neural information
  processing systems, 2008, pp. 1729--1736.

\bibitem{nash1951non}
J.~Nash, Non-cooperative games, Annals of mathematics (1951) 286--295.

\bibitem{foster1999regret}
D.~P. Foster, R.~Vohra, Regret in the on-line decision problem, Games and
  Economic Behavior 29~(1-2) (1999) 7--35.

\bibitem{tammelin2015solving}
O.~Tammelin, N.~Burch, M.~Johanson, M.~Bowling, Solving heads-up limit texas
  hold'em, in: Twenty-Fourth International Joint Conference on Artificial
  Intelligence, 2015.

\bibitem{bowling2015heads}
M.~Bowling, N.~Burch, M.~Johanson, O.~Tammelin, Heads-up limit hold’em poker
  is solved, Science 347~(6218) (2015) 145--149.

\bibitem{brown2019solving}
N.~Brown, T.~Sandholm, Solving imperfect-information games via discounted
  regret minimization, in: Proceedings of the AAAI Conference on Artificial
  Intelligence, Vol.~33, 2019, pp. 1829--1836.

\bibitem{brown2019deep}
N.~Brown, A.~Lerer, S.~Gross, T.~Sandholm, Deep counterfactual regret
  minimization, in: International Conference on Machine Learning, 2019, pp.
  793--802.

\bibitem{johanson2012efficient}
M.~Johanson, N.~Bard, M.~Lanctot, R.~Gibson, M.~Bowling, Efficient nash
  equilibrium approximation through monte carlo counterfactual regret
  minimization, in: Proceedings of the 11th International Conference on
  Autonomous Agents and Multiagent Systems-Volume 2, International Foundation
  for Autonomous Agents and Multiagent Systems, 2012, pp. 837--846.

\bibitem{burch2013cfr}
N.~Burch, M.~Bowling, Cfr-d: Solving imperfect information games using
  decomposition, arXiv preprint arXiv:1303.4441 (2013) 1--15.

\bibitem{lisy2015online}
V.~Lis{\`y}, M.~Lanctot, M.~Bowling, Online monte carlo counterfactual regret
  minimization for search in imperfect information games, in: Proceedings of
  the 2015 international conference on autonomous agents and multiagent
  systems, 2015, pp. 27--36.

\bibitem{brown2015regret}
N.~Brown, T.~Sandholm, Regret-based pruning in extensive-form games, in:
  Advances in Neural Information Processing Systems, 2015, pp. 1972--1980.

\bibitem{moravvcik2017deepstack}
M.~Morav{\v{c}}{\'\i}k, M.~Schmid, N.~Burch, V.~Lis{\`y}, D.~Morrill, N.~Bard,
  T.~Davis, K.~Waugh, M.~Johanson, M.~Bowling, Deepstack: Expert-level
  artificial intelligence in heads-up no-limit poker, Science 356~(6337) (2017)
  508--513.

\bibitem{Brown2017Superhuman}
N.~Brown, T.~Sandholm, Superhuman ai for heads-up no-limit poker: Libratus
  beats top professionals., Science 359~(6374) (2017) 1733.

\bibitem{brown2019superhuman}
N.~Brown, T.~Sandholm, Superhuman ai for multiplayer poker, Science 365~(6456)
  (2019) 885--890.

\bibitem{jackson2016compact}
E.~G. Jackson, Compact cfr, in: Workshops at the Thirtieth AAAI Conference on
  Artificial Intelligence, 2016.

\bibitem{jackson2017targeted}
E.~G. Jackson, Targeted cfr, in: Workshops at the Thirty-First AAAI Conference
  on Artificial Intelligence, 2017.

\bibitem{li2018double}
H.~Li, K.~Hu, Z.~Ge, T.~Jiang, Y.~Qi, L.~Song, Double neural counterfactual
  regret minimization, arXiv preprint arXiv:1812.10607.

\bibitem{steinberger2019single}
E.~Steinberger, Single deep counterfactual regret minimization, arXiv preprint
  arXiv:1901.07621.

\bibitem{zhou2018lazy}
Y.~Zhou, T.~Ren, J.~Li, D.~Yan, J.~Zhu, Lazy-cfr: fast and near optimal regret
  minimization for extensive games with imperfect information, arXiv preprint
  arXiv:1810.04433.

\bibitem{schmid2019variance}
M.~Schmid, N.~Burch, M.~Lanctot, M.~Moravcik, R.~Kadlec, M.~Bowling, Variance
  reduction in monte carlo counterfactual regret minimization (vr-mccfr) for
  extensive form games using baselines, in: Proceedings of the AAAI Conference
  on Artificial Intelligence, Vol.~33, 2019, pp. 2157--2164.

\bibitem{mnih2015human}
V.~Mnih, K.~Kavukcuoglu, D.~Silver, A.~A. Rusu, J.~Veness, M.~G. Bellemare,
  A.~Graves, M.~Riedmiller, A.~K. Fidjeland, G.~Ostrovski, et~al., Human-level
  control through deep reinforcement learning, Nature 518~(7540) (2015) 529.

\end{thebibliography}

\clearpage
\section*{Appendix}
\label{proof}

\begin{center}
	\textbf{Proof of Theorem 1}
\end{center}

Consider the weighted sequence of iterations Consider the weighted sequence of iterations $\sigma^{\prime 1}, \ldots, \sigma^{\prime T}$, where $\sigma^{\prime t}$ is identical to $\sigma^t$, but weighted by $w_t$. The regret of action $a$ in information set $I$ on iteration $t$ of this new sequence is $R^{'t}(I,a)$.


\textbf{Theorem 1}  Assume that the number of the iteration $T$ of the RLCFR, which is played in a two-player zero-sum game. Then the weighted average strategy profile is a 
$2\frac{\Delta |\mathcal{I}| \sqrt{|A|}} {\sqrt{T}}$-Nash equilibrium.

The proof is provided in the following, which combines the the proof for the vanilla CFR \cite{zinkevich2008regret}, CFR+ \cite{tammelin2015solving,bowling2015heads} and the discount CFR \cite{brown2019solving}.

\textit{Proof}. The lowest amount of the instant regret on any iteration is $-\Delta$. Consider the weighted sequence of iterations $\sigma^{\prime 1}, \ldots, \sigma^{\prime T}$, where $\sigma^{\prime t}$ is identical to $\sigma^t$, but the weight $w_{a,t}=\prod_{i=t}^{T-1} i =\frac{(T-1)!}{t!}$ rather than $w_{a,t}=(1, e^\alpha, t, (\frac{t}{t+1})^\gamma, 1, 1, 1)$. $R^{\prime t}(I, a)$ is the regret of the action $a$ on the information set $I$ at the iteration $t$, for this new sequence. 

Here, we have made a contraction of $w_{a,t}$. $w_{a,t}=(1, e^\alpha, t, (\frac{t}{t+1})^\gamma, 1, 1, 1)\leq (t, t, t, t, t, t, t)$. In the RLCFR, it needs to be noted that we select the way of strategy updating with the iteration increases. Specifically, the weight varies as the number of iteration increases. About $w_{a,t}=(1, e^\alpha, t, (\frac{t}{t+1})^\gamma, 1, 1, 1)$, it stands for seven kinds of weights, which comes from seven different ways of strategy updating. Moreover, one is selected to be adopted from seven kinds of weights on each iteration in the paper. 

In addition, for the regret matching \cite{Cesa2006Prediction}, which proves that if $\sum_{t=1}^{\infty} w_{t}=\infty$, then the weighted average regret that defined as $R_{i}^{w, T}=\max _{a \in A} \frac{\sum_{t=1}^{T}\left(w_{t} r^{t}(a)\right)}{\sum_{t=1}^{T} w^{t} }$ can be bounded by the $R_{i}^{w, T} \leq \frac{\Delta \sqrt{|A|} \sqrt{\sum_{t=1}^{T} w_{t}^{2}}}{\sum_{t=1}^{T} w_{t}}$.

We can find that $R^t(I,a) \leq \Delta\sqrt{|A|} \sqrt{T}$ for the player $i$' action $a$ on the information set $I$, from the lemma 3. We can use the Lemma 1, which uses the  weight $w_{\alpha, t}$ for the iteration $t$ with $B= \Delta\sqrt{|A|} \sqrt{T}$ and $C=0$. It means that $R^{\prime t}(I, a) \leq w_T(B-C) \leq \Delta\sqrt{|A|} \sqrt{T} $ from the Lemma 1.  Furthermore, we get the weighted average regret is at most $\Delta\sqrt{|A|} \sqrt{T}$ from the Lemma 3. Since for the information set,  $\left|\mathcal{I}_{1}\right|+\left|\mathcal{I}_{2}\right|=|\mathcal{I}|$, and as $T \rightarrow \infty$, $\frac{R_i^T}{T} \rightarrow 0 $. Therefore, in the two-player zero-sum game, this weighted average strageties form a $2\frac{\Delta |\mathcal{I}| \sqrt{|A|}} {\sqrt{T}}$-Nash equilibrium.

\textbf{Lemma 1}. Call a sequence $x_{1}, \ldots, x_{T}$ of bounded real values BC-plausible if $B>0, C \leq 0, \sum_{t=1}^{i} x_{t} \geq C$ for
all $i,$ and $\sum_{t=1}^{T} x_{t} \leq B .$ For any $B C$ -plausible sequence and any sequence of non-decreasing weights $w_{t} \geq 0$,  $\sum_{t=1}^{T}\left(w_{t} x_{t}\right) \leq w_{T}(B-C)$.

\textbf{Lemma 2}. Given a group of actions $A$ and any sequence of rewards $v^{t}$, such that $\left|v^{t}(a)-v^{t}(b)\right| \leq \Delta$ for all $t$ and all $a, b \in A,$ after conducting a set of strategies decided by the regret matching, however applying the regret-like value $Q^{t}(a)$ instead of $R^{t}(a), Q^{T}(a) \leq \Delta \sqrt{|A| T}$ for all $a \in A$.

\textit{Proof}. This Lemma is closely resembling Lemma 1, which are both from \cite{tammelin2015solving}, thus here we donot give the detailed proof of these two lemmas. 

\textbf{Lemma 3}. Suppose the player $i$ conducts $T$ iterations of the ECFR, then the weighted regret for the player $i$ is at most $\Delta\left|\mathcal{I}_{i}\right| \sqrt{|A|} \sqrt{T}$, and the weighted average regret for the player $i$ is at most $\frac{\Delta\left|\mathcal{I}_{i}\right| \sqrt{|A|}}{ \sqrt{T}}$.

\textit{Proof}. The weight of the iteration $t<T$ is $w_t=\prod_{i=t}^{T-1} i$. Therefore, for all iteration $t$, $\sum_{t=1}^{T} w_{t}^{2} \leq \sum_{t=1}^{T} [\frac{(T-1)!}{t!}]^2  \leq T[(T-1)!]^2$. In addition, $\sum_{t=1}^{T} w_{t} \geq (T-1)!$.

Through $R_{i}^{w, T} \leq \frac{\Delta \sqrt{|A|} \sqrt{\sum_{t=1}^{T} w_{t}^{2}}}{\sum_{t=1}^{T} w_{t}}$ and the Lemma 2, we can find that $ Q_{i}^{w, T} \leq \frac{\Delta \sqrt{|A|} \sqrt{\sum_{t=1}^{T} w_{t}^{2}}}{\sum_{t=1}^{T} w_{t}}\leq \Delta \sqrt{|A|} \sqrt{T}$. Due to $ R_i^T\leq \sum_{I \in \mathcal{I}_i} R^T(I)$ \cite{Cesa2006Prediction}, we can find that $ Q_{i}^{w, T} \leq \Delta\left|\mathcal{I}_{i}\right| \sqrt{|A|} \sqrt{T}$. Since $R_i^{w,T} \leq Q_i^{w,T}$, thus $R_i^{w,T} \leq \Delta\left|\mathcal{I}_{i}\right| \sqrt{|A|} \sqrt{T}$.

\end{document}